\documentclass[conference]{IEEEtran}
\IEEEoverridecommandlockouts
\usepackage{multirow}
\usepackage{multicol}
\usepackage[utf8]{inputenc} 
\usepackage{kotex} 

\usepackage{hyperref}
\usepackage{cite}
\usepackage{amsmath,amssymb,amsfonts}
\usepackage{algorithm}
\usepackage{algorithmicx}
\usepackage{algpseudocode}
\usepackage{graphicx}
\usepackage{textcomp}
\usepackage{xcolor}
\usepackage{cite}
\usepackage{xspace}
\usepackage{soul}
\usepackage{kotex}
\usepackage{booktabs} 
\usepackage{hyperref}
\definecolor{linkcolour}{rgb}{0,0.2,0.6}
\definecolor{xgreen}{rgb}{0.2,0.6,0.0}
\definecolor{xred}{rgb}{0.7,0.1,0.0}
\hypersetup{colorlinks,breaklinks,citecolor=red!50, urlcolor=linkcolour, linkcolor=xgreen}

\usepackage{amsmath}
\usepackage{color}
\usepackage{array}
\usepackage{stfloats}
\usepackage{amsthm} 
\usepackage{amssymb}
\usepackage{float}
\usepackage{nccmath}
\usepackage{graphicx}
\usepackage{setspace}
\usepackage[normalem]{ulem}
\usepackage{booktabs}
\usepackage{subcaption}
\usepackage{mwe}
\usepackage{cite}
\usepackage{graphicx}
\usepackage[export]{adjustbox}
\usepackage{amsmath,amssymb,amsfonts}
\usepackage{graphicx}
\usepackage{textcomp}
\usepackage{xcolor}
\usepackage[utf8]{inputenc} 
\usepackage{kotex} 

\usepackage{algorithm} 
\usepackage{algpseudocode}
\usepackage{grffile}

\usepackage{kotex}
\newcommand{\RN}[1]{%
  \textup{\uppercase\expandafter{\romannumeral#1}}%
}

\def\BibTeX{{\rm B\kern-.05em{\sc i\kern-.025em b}\kern-.08em
    T\kern-.1667em\lower.7ex\hbox{E}\kern-.125emX}}
\begin{document}
\bstctlcite{IEEEexample:BSTcontrol}
\title{Tutorial on Course-of-Action (COA) Attack Search Methods in Computer Networks}

\author{
\IEEEauthorblockN{Seok Bin Son$^\dag$, Soohyun Park$^\dag$, Haemin Lee$^\dag$, Joongheon Kim$^\dag$, Soyi Jung$^\ddag$, and Donghwa Kim$^\S$}
\IEEEauthorblockA{
$^\dag$Department of Electrical and Computer Engineering, Korea University, Seoul, Republic of Korea\\
$^\ddag$School of Software, Hallym University, Chuncheon, Republic of Korea\\
$^\S$Agency for Defense Development (ADD), Seoul, Republic of Korea\\
\texttt{\{lydiasb,soohyun828,haemin2,joongheon\}@korea.ac.kr}, 
\texttt{sjung@hallym.ac.kr}, 
\texttt{dhkim@add.re.kr}}
}

\maketitle

\begin{abstract}
 In the literature of modern network security research, deriving effective and efficient course-of-action (COA) attach search methods are of interests in industry and academia. As the network size grows, the traditional COA attack search methods can suffer from the limitations to computing and communication resources. Therefore, various methods have been developed to solve these problems, and reinforcement learning (RL)-based intelligent algorithms are one of the most effective solutions. Therefore, we review the RL-based COA attack search methods for network attack scenarios in terms of the trends and their contributions.
\end{abstract}

\section{Introduction}
With the development of large-scale complex networks based on industrial technological development and the possibility of numerous cyber threats, cyber security has become one of the most important areas of research areas. A penetration testing is a cyber security approach that involves testing the network environment in order to assess system security or identify vulnerabilities. In this paper, we define the penetration testing as the attack of a network's course of action (COA) that may be used to strategically optimize decision making in a variety of network environments to ensure system security. 

\begin{figure}[t]
    \begin{center}
        \includegraphics[width=0.95\linewidth]{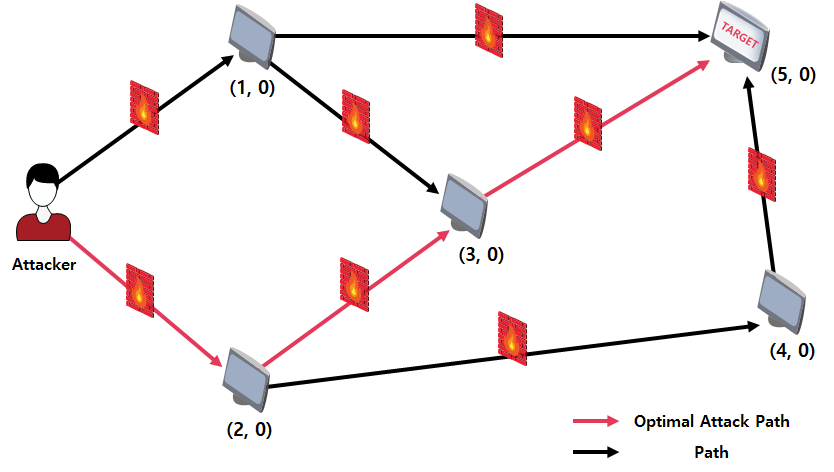}
    \end{center}
    \caption{A reference network COA attack scenario.}
    \label{fig:Network_Attack_Scenario}
\end{figure}

There are two types of COA attack search techniques, i.e., \textit{passive} and \textit{automated}. The traditional passive COA attack search method requires the participation and direction of security experts, making it inefficient in terms of time and cost.
Therefore, various automated algorithms have been designed to search COA attack automatically such as Attack Tree~\cite{Attack_Tree}, Attack Graph~\cite{Attack_Graph}, and Game Theory~\cite{Game_Theory}. These existing algorithms, on the other hand, have drawbacks due to the dependence on data learning or inability to perform well in uncertain network environments. For the given problems, it has been proved that reinforcement learning (RL)-based algorithm~\cite{RL} can overcome data drawbacks by learning and determining the best policy in a dynamic context~\cite{pieee202105park,9647911,iotj20kwon,tvt202106jung,tvt202108jung,jsac201806choi}. As a result, it is possible to efficiently discover the best attack path in a given network and check the system's security, when the RL algorithm is used to COA attack for cyber, as shown in Fig
.~\ref{fig:Network_Attack_Scenario}. As a result, several researches have been conducted to automatically search COA attack and enhance reliability using the RL algorithms. 
Finally, the main purpose of this paper is to present and summarize several important researches on COA cyber attack using RL algorithms.

The rest of this paper is organized as follows. Sec.~\ref{sec:sec2} presents the definition of RL-based algorithms. Sec.~\ref{sec:sec3} also presents various research results on COA attack using the RL-based algorithms. In addition, Sec.~\ref{sec:sec4} presents potential emerging research directions. Finally, Sec.~\ref{sec:sec5} concludes this paper and presents future work directions.

\section{RL Algorithms used in COA Attack Search Methods}\label{sec:sec2}
In this section, we summarize the RL algorithms used in COA attack search. In Sec.~\ref{sec:sec2:A}, the definition of RL-based algorithms is described. In Sec.~\ref{sec:sec2:B} the RL algorithms that are most commonly used in the COA attack search methods are described.

\subsection{Definition of RL-based Algorithms}\label{sec:sec2:A}

\begin{figure}[t]
    \begin{center}
        \includegraphics[width=0.95\linewidth]{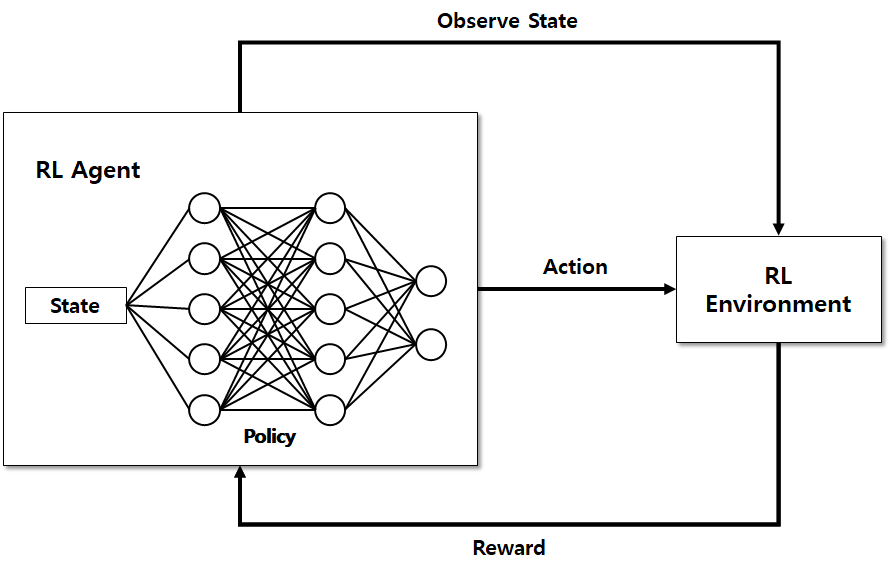}
    \end{center}
    \caption{RL algorithm.}
    \label{fig:RL}
\end{figure}

The RL algorithm is an algorithm that makes decisions according to the fundamental concepts of Markov decision process (MDP), which can be expressed as illustrated in Fig.~\ref{fig:RL}~\cite{twc201912choi}. The MPD repeats the following four processes in order to conduct optimal decision-making, i.e., (i) state observation, (ii) behavioral decision, (iii) state transition, and (iv) next state and immediate compensation. That is, the agent observes the environment's state information and uses the policy to probabilistically determine what action to take for maximizing expected return where the return isdefined as the summation of the rewards by sequential action taking. 
When the action is completed, the corresponding reward is given, and the process is repeated until the system is terminated. The agent iteratively goes through these steps again and again, for making decisions that maximize the total reward. The RL algorithm that operates in this way has been actively studied in various fields such as autonomous driving and quantum deep learning\cite{RL_AIMLAB_1, RL_AIMLAB_2,RL_AIMLAB_3}.

\subsection{Classification of RL Algorithms}\label{sec:sec2:B}
The RL algorithms are also widely used in various cyber security fields, among which COA attack method is used. There are various algorithms those are designed using RL-based methods, however the algorithms used mainly in recent COA attack searches are as follows, i.e., partially observable Markov decision processes (POMDP)~\cite{POMDP}, Q-learning~\cite{Q-learning}, deep Q-network (DQN)~\cite{DQN}, advanced actor critic (A2C)~\cite{A2C}, and proximal policy optimization (PPO)~\cite{PPO}.
More details about these algorithms are as follows.

\begin{itemize}
    \item\textit{POMDP} is an RL algorithm that makes decisions considering the situation that the agent should communicate with an uncertain environment.

    \item\textit{Q-learning} is a model-free RL algorithm that determines the behavior in a given situation with the highest Q-value.

    \item\textit{DQN} is an RL algorithm that uses the deep neural network to construct a Q-function that represents Q-value in Q-learning.

    \item\textit{A2C} is a policy-based RL algorithm that uses a critic network to update value functions and an actor network to change parameters.

    \item\textit{PPO} is also a policy-based RL algorithm that determines optimal behavior probability values by approximating policy neural networks.

\end{itemize}

\section{Trends in RL-based COA Cyber Attack Search Methods}\label{sec:sec3}

\subsection{POMDP}
In the research results using POMDP~\cite{POMDP1, POMDP2}, the information about network components (e.g., network topology) is considered important, so that the fusion of scanning and exploit among the actions of COA attack search methods can be intelligent. Furthermore, according to the research results in~\cite{POMDP3, POMDP4}, they propose intelligent automated penetration testing systems (IAPTS), therefore, the COA attack search methods can be performed intelligently and autonomously using the POMDP-based algorithms for enhancing accuracy, saving time and money, and also increasing efficiency.

\subsection{Q-Learning}
In the research results using Q-Learning~\cite{Q-learning2}, it is able to explore the optimal COA attack paths within the network, even if not all information exists for the network to which the COA attack search methods will be applied. Moreover, in a research~\cite{Q-learning4} that proposed a framework that integrates Q-Learning algorithms with ontology-based belief desire intention (BDI), COA attack using that framework was particularly adaptable because the optimal attack path could be found in just a few attempts even when there was no information about the network. In addition, research~\cite{Q-learning1, Q-learning3} using various algorithms (e.g., Random, Greedy, Q-Learning, DQN, and etc) revealed that the Q-Learning algorithm performed best in the COA attack search methods.

\subsection{DQN}
In the research results using the combination of DQN algorithm and intelligence preparation of the battlefield (IPB)~\cite{DQN1}, the optimal attack path was better found by using the cyber topography, and the reward acquisition was higher. In addition, the research result in~\cite{DQN2} based on DQN and common vulnerability scoring system (CVSS) score information can be used to find the best possible attack paths in a given environment. 
In the research~\cite{DQN3} using NDSPI-DQN (i.e., noisy nets, dueling network architectures, soft Q-learning, prioritized experience replay, and intrinsic curiosity module in DQN), the algorithm developed from DQN, was used to improve the navigation ability on attack paths, and its trial and error cost of decision makers could be reduced through spatial vector separation. Furthermore, the research result in~\cite{DQN4} that presented hierarchical agent deep reinforcement learning (HA-DRL), another algorithm developed in DQN, said that large-scale discrete action space that occurs when COA attack method is performed can be efficiently handled, and the corresponding performance is improved compared to the case with a single DQN.

\subsection{A2C}
In the research results using a combination of A2C algorithms and double agent environments~\cite{A2C1} , the COA attack search methods performed well in identifying information used to perform COA attack research methods (e.g., assessment indicators used for risk assessment of a given network environment, services used to utilize data, and etc).

\subsection{PPO}

In the research results that combine the PPO algorithm with random network distillation (RND) to select as the agent of COA attack search methods~\cite{PPO1}.
The proposed methods learn with sparse environment rewards; as well as propose to use multi-objective Markov decision process (MOMDP) to perform automatic COA attack search methods. In addition, the research result is proposed to generate various behavioral agents through the Chebyshev deformation critique to find various attack steps that balance the different purposes of the COA attack search methods.

\section{Discussions}\label{sec:sec4}
The RL-based algorithm for COA attack search performs effectively regardless of the prior knowledge of the network environment. Based on these advantages, many research results have been published using the RL algorithms as shown in Sec.~\ref{sec:sec3}. However, there is a disadvantage that the performance difference occurs depending on the size of the network, when the RL algorithm is used for COA attack search~\cite{DQN3}. The COA attack search method is an attack vector in which an attack is ordered among various attack paths. In this case, the learning of the RL agent decreases, as the size of the network increases. This problem occurs based on the following reasons.

\begin{itemize}
    \item As the size of the network increases, the size of the action spaces increases, so that the agent of the RL-based algorithm becomes difficult to explore.

    \item The number of the hosts with positive values in a network environment is only a few. As the network size increases, it is difficult to converge because the reward occurs sparsely. Therefore, the reward that the RL agent can obtain becomes scarce, and thus, it becomes difficult to converge in the RL-based algorithm~\cite{Reason2}.
\end{itemize}

\section{Concluding Remarks}\label{sec:sec5}
As the interest in security and privacy has increased, many research results have been conducted on the COA attack search method as a preemptive response method to check the security of the network. Based on this reason, various autonomous COA attack search methods have been introduced to overcome the limitations to the traditional methods of COA attack navigation in passive ways that have been inefficient in terms of time and cost. However, these algorithms have the disadvantage of being difficult to apply in an uncertain environment of the network. Based on this fact, it is clear that the RL-based algorithm can be very useful to find the optimal attack path even in uncertain network environments. As such, COA attack search methods are developing from passive analysis technologies to automated analysis technologies. Among them, research results are applying RL algorithms to COA attack search are drawing attention. Therefore, in this paper, we describe the trend of research applying RL algorithms to COA attack search and show that the use of RL algorithms can improve performance.

\section*{Acknowledgment}
This work was supported by the Agency for Defense Development under the contract UI210009XD. J. Kim is a corresponding author of this paper.

\bibliographystyle{IEEEtran}
\bibliography{ref_aimlab}

\end{document}